# Toward Improving Health Literacy in Patient Education Materials with Neural Machine Translation Models

David Oniani[1], Sreekanth Sreekumar[1], Renuk DeAlmeida[2], Dinuk DeAlmeida[2], Vivian Hui, RN, PhD[3], Young Ji Lee, RN, PhD[4,6], Yiye Zhang, PhD[5], Leming Zhou, PhD[1], Yanshan Wang, PhD[1,6,7]

[1]Department of Health Information Management, University of Pittsburgh, Pittsburgh, PA; [2]North Allegheny Intermediate High School, Pittsburgh, PA; [3]School of Nursing, The Hong Kong Polytechnic University; [4]School of Nursing, University of Pittsburgh, Pittsburgh, PA; [5]Division of Health Informatics, Weill Cornell Medicine, New York, NY; [6]Department of Biomedical Informatics, University of Pittsburgh, Pittsburgh, PA; [7]Intelligent Systems Program, University of Pittsburgh, Pittsburgh, PA

**Abstract**

*Health literacy is the central focus of Healthy People 2030, the fifth iteration of the U.S. national goals and objectives. People with low health literacy usually have trouble understanding health information, following post-visit instructions, and using prescriptions, which results in worse health outcomes and serious health disparities. In this study, we propose to leverage natural language processing techniques to improve health literacy in patient education materials by automatically translating illiterate languages in a given sentence. We scraped patient education materials from four online health information websites: MedlinePlus.gov, Drugs.com, Mayoclinic.org and Reddit.com. We trained and tested the state-of-the-art neural machine translation (NMT) models on a silver standard training dataset and a gold standard testing dataset, respectively. The experimental results showed that the Bidirectional Long Short-Term Memory (BiLSTM) NMT model outperformed Bidirectional Encoder Representations from Transformers (BERT)-based NMT models. We also verified the effectiveness of NMT models in translating health illiterate languages by comparing the ratio of health illiterate language in the sentence. The proposed NMT models were able to identify the correct complicated words and simplify into layman language while at the same time the models suffer from sentence completeness, fluency, readability, and have difficulty in translating certain medical terms.*

**Introduction**

Healthy People 2030 is the fifth iteration of U.S. national goals and objectives aimed at promoting health and disease prevention.[1] Health literacy is a central focus of Healthy People 2030 and is one of five overarching goals since it is a critical component to address health disparities, to achieve health equity, and eventually to improve the health and well-being of all[2]. As defined by Healthy People 2030, Personal Health Literacy (PHL) is "the degree to which individuals have the ability to find, understand, and use information and services to inform health-related decisions and actions for themselves and others."[3] The new PHL definition focuses on consumers' ability to use the patient education materials and make well-informed decisions. According to the National Assessment for Adult Literacy (NAAL), around 80 million Americans, which is over 36% of adult Americans, were deemed to be at or below the basic level of health literacy.[4] The NAAL results showed that socioeconomic factors, education level, race and age are major factors that influence health literacy. These factors may also work together to further worsen disparities in healthcare as certain groups are more in need of healthcare and have the burden of the higher costs associated with it. In particular, more than 58% of African Americans had basic or below basic health literacy[5], and over 74% of Spanish-speaking patients have less-than-adequate health literacy as compared to 7% of English-speaking patients[6]. On average, adults aged 65 and older have lower health literacy than adults under the age of 65, and older adults with cognitive impairment have significantly lower health literacy[7].

According to a recent study,[8] 75% of patient education materials are written at high school or college level, while the general American adult read at an eighth-grade level. Patients with low health literacy have had trouble understanding health information, following post-visit instructions, managing chronic conditions, reading prescription labels, and using and administering medication. These all result in worse health outcomes and serious health disparities as patients are not able to manage their own health properly. Patients who have difficulty understanding those materials usually search for relevant health information on the Internet. Thus, the Internet has become an important source for these patients to obtain health information. However, similar to the patient education materials distributed through hospitals and clinics, many studies have found poor readability and illiterate language in the content of online health information.[9,10] To improve the health literacy in patient education materials, manually rewriting the content is labor intensive and infeasible. Therefore, we propose to leverage natural language processing (NLP) techniques to address the issue of frequent illiterate language, which is difficult to understand for people with low health literacy, and translate the illiterate language into more literate and understandable language.

**Related Work**

In the general NLP domain, translating language into a more understandable format has been studied in the area of sentence simplification. The most conventional approach is based on rewriting a sentence by substituting rare words with more common words or phrases and reducing the complexity of grammar structure. For example, Chandrasekar et al.[11] used syntactic rules to manually split a complex sentence into simpler sentences. WordNet[12] was later used to find synonyms and phrases to substitute difficult words in a sentence.[13] However, these conventional approaches are mostly based on rules that are tedious and need lots of human manual effort. In addition, substituting complex words may lead to grammatically incorrect sentences, particularly if the complex words are verbs.[13] As machine and deep learning models achieved state-of-the-art performance in many NLP tasks, they have been adopted for automated sentence simplification. Statistical machine translation methods have been used to generate simple aligned sentences from the original complex sentences either as a stand-alone approach or in a hybrid approach.[14,15] Zhang and Lapata[16] used Recurrent Neural Network (RNN) and Long Short Term Memory (LSTM) in combination with reinforcement learning to generate simple sentences. However, most approaches in the literature of general NLP domain are not studied in the healthcare domain due to the lack of labeled corpus for medical language.

The need to simplify medical language was identified two decades ago by the National Institute of Health (NIH) and Centers for Disease Control and Prevention (CDC). Several guidelines have been developed to train health providers for writing simple medical texts and communicating with patients using plain language, for example, the Plain Language Initiative or the California Health Literacy initiative. However, these guidelines were not ideal to be used in practice due to its complexity. A study by the American Medical Informatics Association's Consumer Health Informatics Working Group in the report[16,17] reviewed existing theory and informatics tools in consumer health communication and found that these tools did not meet consumers' need to understand the retrieved health information. Readability of health information has been studied and assessed in follow-up studies[18] and a rule-based simplification tool was developed to simplify health information[19]. Kauchak et al.[19,20] focused on the assessment of grammar difficulty and methods for grammar simplification, which is an important aspect of improving readability of health information. However, these methods are based on manual rules to substitute different medical terms, which are time-consuming to develop and lack generalizability.

**Materials and Methods**

In this study, we tested the state-of-the-art neural machine translation models (NMTs) for translating complex and illiterate language into simple and literate language in patient education materials. Since there is no public labeled dataset for this purpose, we firstly created a corpus to train the NMT models. Then we tested two NMT methods to automatically translate the illiterate health language. In the following subsections, we described the details of dataset creation and NMT models.

*Dataset Curation*

We used four online health information websites to create the corpus for training NMT models: MedlinePlus.gov, Drugs.com, Mayoclinic.org and Reddit.com. We utilized Python web crawler tools to scrape data from Mayoclinc.org and Drugs.com and leveraged publicly available API's for MedlinePlus.gov and Reddit.com. For Mayoclinic.org, we captured the data in domains like diseases and conditions, symptoms, tests and procedures, drugs and supplements, etc. For MedlinePlus.gov, we scraped the available health information and downloaded the XML files. For Reddit.com, we scraped data from the popular medically pertaining subreddits such as r/medicine, r/medical_news, and r/medical. For Drugs.com, we crawled all the drug descriptions, side effects, dosage, condition overview, risks, symptoms, diagnosis, and treatment. The data scraping was conducted in May, 2022.

After scraping the data from the four online resources, we built a corpus of sentence snippets with health illiterate languages. We used a dictionary, the CDC Plain Language Thesaurus, to identify illiterate language. The CDC Plain Language Thesaurus offers a mapping for commonly health illiterate words or phrases that are often used by physicians and medical practitioners to its corresponding health literate form. We split the patient education materials into sentence snippets, pre-processed the data after removal of short snippets (such as titles) and snippets with hyperlinks, and obtained a set of sentence snippets that contain at least one of the illiterate words based on the CDC Plain Language Thesaurus. Eventually, we obtained a total of 285,348 sentences, including 176,954 sentences from Mayoclinic.org, 89,492 from Drugs.com, 1,937 from MedlinePlus.gov 16,965 from Reddit.com.

*Silver Standard Dataset Creation*

Due to the facts that 1) NMTs usually need a large dataset for training, 2) to the best of our knowledge, there is no public dataset for training models to translate health illiterate language, and 3) manually creating a gold standard dataset is time-consuming, we created a silver standard dataset to train the NMTs for translating the health illiterate language. We used the illiterate-literate word or phrase mapping in the CDC Plain Language Thesaurus to create the silver standard dataset. Specifically, we randomly substitute one health illiterate word in each sentence with a literate word or phrase according to the mapping. The sentences in our dataset may contain more than one health illiterate word. Based on our observation, substituting multiple illiterate words usually makes the sentence nonsense. That will introduce noise when training the model and may generate nonsense sentences during testing. Therefore, we only substituted one illiterate word for each sentence in our silver standard dataset. Having said that, substituting one illiterate word may still result in semantically and grammatically incorrect sentences. Thus, we applied an extra grammar correction tool, named T5 grammar correction based on a pre-trained language model, to automatically correct grammar and polish the sentence. Finally, we obtained a silver standard dataset of sentence pairs with illiterate words (called *source* in machine translation) and limited literate words (called *target* in machine translation). The silver standard dataset was split into training with 245,335 sentence pairs and validation with 40,000 pairs to train and validate every NMT model. In order to test the NMT models, we also built a gold standard dataset of 497 sentence snippets with manually translated results of all illiterate languages in the source.

*Neural Machine Translation Models*

We trained and tested two NMT methods[21] to automatically translate health illiterate language. Compared to the rule-based and statistical machine translation (SMT) approaches, the NMT system has achieved remarkable performance and has been widely adopted as a reliable NLP tool for machine translation. The first NMT method is based on a Bidirectional Long Short-Term Memory (BiLSTM) architecture[21]. As illustrated in Figure 1, BiLSTMs utilize a sequence processing model which includes an LSTM for taking the input in the forward direction, and another LSTM taking the input in the backward direction. BiLSTMs were chosen for our model's architecture because it provides additional training since it traverses the input data twice which leads to higher quality translations. Our model is trained using source files which consist of health illiterate and target files with the source files translated to become health literate. The model learns from the translations in the target files to translate new health illiterate sentences. We used the implementation of BiLSTM provided by OpenNMT package[1].

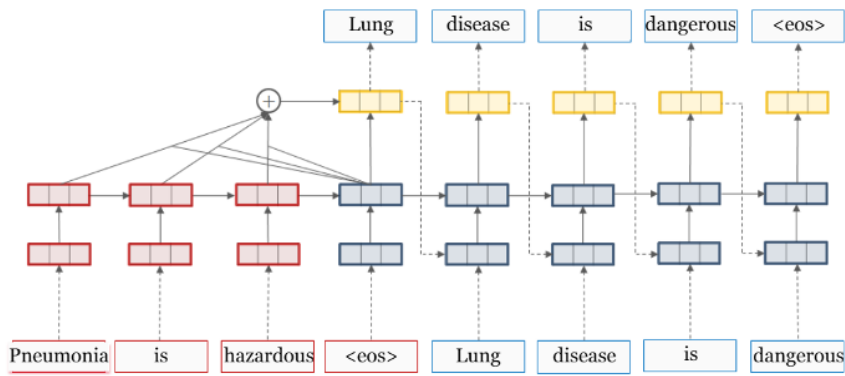

**Figure 1**. Architecture of the BiLSTM model for translating the health illiterate language.

The second NMT method is based on the Transformer architecture.[22] Transformers were initially introduced in the context of machine translation in order to tackle two major challenges: 1) avoiding recursion in order to allow parallel computation and reduce training time; and 2) reducing drops in performance due to long dependencies (i.e., vanishing gradient in RNNs). Using transformer-based NMT methods will address the above challenges and thus be used in this study to tackle the health illiterate language translation task. Additionally, transformer-based models have shown the state-of-the-art performance in a number of tasks including text generation[23] and image recognition[24].

---
[1] https://opennmt.net/

The architecture is shown in Figure 2. Models were composed of a single encoder and a single decoder, each with 12 attention blocks. The encoder contains encoder blocks that process the input in a sequence, one block after another. The decoder contains decoder blocks that, similar to the encoder, processes the data one block after another. And while the encoder processes the input data, the decoder processes the data produced by the encoder (i.e., the encoder output). Every one of the 12 encoder blocks is the same and contains a multi-head attention, add (a residual connection that adds the input of each layer to the output), layer normalization operation, feed forward layer, and another add & norm. All 12 decoder blocks are also architecturally the same and contain two sets of multi-head attention, add, and layer normalization operations followed by a feed forward layer and another add & norm. We used pre-trained transformed-based encoder-decoder neural language models (PLMs). Transformer-based PLMs have achieved state-of-the-art performance in data-poor NLP tasks by transferring the prior knowledge learned from a large unlabeled corporus in an unsupervised way. We evaluate three transformer-based encoder-decoder PLMs - Bidirectional Encoder Representations from Transformers (BERT)[25], Bidirectional Encoder Representations from Transformers for Biomedical Text Mining (BioBERT)[26], and Bio + Clinical BERT (BioClinicalBERT)[27].

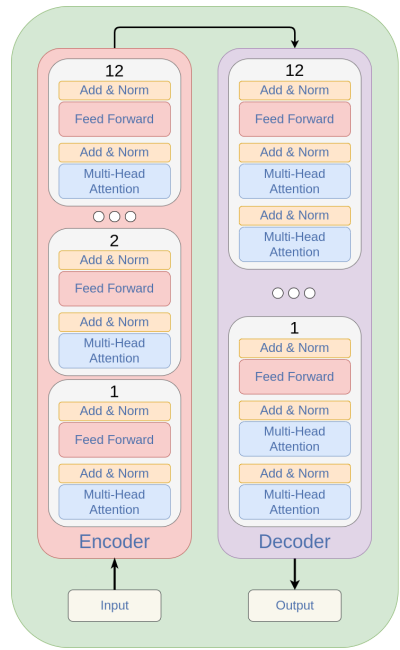

**Figure 2**. Architecture of the Transformer-based NMT model.

Transformer-based models initially undergo the unsupervised training, where part of the input is masked and the transformer attempts to approximate the masked part. All language transformers used in our experiments were already pre-trained in the unsupervised manner. We have fine-tuned the pre-trained models on the downstream task - illiterate language translation.

*Evaluation Metrics*

We have used the BLEU (BiLingual Evaluation Understudy) score[28] for automatically evaluating the performance of the neural machine translation models. The BLEU score is typically a number between 0 and 100 (but sometimes can also be represented on a 0 - 1 scale). A score of 0 means that the machine-translated output has no overlap with the reference translation and hence, the translation is of a low quality. A score of 100, however, means that there is a perfect overlap between the machine-translated output and the reference translation, which would mean that the translation is of a perfect quality. It should also be noted that achieving the perfect score of 100 is a very challenging task even for human translators. Readers could refer to the guideline in Table 1 for the BLEU score interpretation.[2]

**Table 1**. The interpretation guidelines for BLEU scores.[3]

| BLEU Score | Interpretation |
|---|---|
| < 10 | Almost useless |
| 10 - 19 | Hard to get the gist |
| 20 - 29 | The gist is clear, but has significant grammatical errors |
| 30 - 40 | Understandable to good translations |
| 40 - 50 | High quality translations |
| 50 - 60 | Very high quality, adequate, and fluent translations |
| > 60 | Quality often better than human |

Additionally, we measured the Health Illiterate Rate (HIR) in each sentence snippet by the following definition:

---

[2] https://cloud.google.com/translate/automl/docs/evaluate#interpretation

$$HIR(sentence) = count(illiterate\ words) / count(total\ words)$$

where the list of illiterate words was derived based on the CDC Plain Language Thesaurus. We calculated the HIR for the original corpus in training, silver standard in training, gold standard in testing, and the output generated by OpenNMT, BERT, BioBERT, and BioClinicalBERT based approaches, and reported the outcomes separately for four online sources.

**Results**

Table 2 lists BLEU scores for the NMT methods based on BiLSTM, BERT, BioBERT, and BioClinicalBERT. As shown in Table 2, the BiLSTM model has outperformed all other models, with the mean BLEU score of 41.578 and 25th, 50th (i.e., the median), and 75th percentile BLEU score values of 25.363, 40.221, and 56.140, respectively. The second best model was BERT, which did better than BioBERT and BioClinicalBERT in all metrics except for the 50th percentile BLEU score, where BioBERT marginally outperformed BERT. The mean BLEU score for the BERT model was 33.582, with the 25th, 50th, and 75th percentile scores of 19.012, 31.233, and 47.355, respectively. BioBERT came next with the mean BLEU score of 33.278 and the 25th, 50th, and 75th percentile values of 18.710, 31.860, and 47.288 respectively. BioClinicalBERT showed the worst performance among all models in our experiments. The mean BLEU score was 31.191 and the 25th, 50th, and 75th percentile BLEU scores were of 18.141, 30.093, and 43.280 respectively.

Table 2. BLEU Scores for OpenNMT-, BERT-, BioBERT-, and BioClinicalBERT-based NMT approaches.

| Model | 25th Percentile | 50th Percentile | 75th Percentile | Mean |
| --- | --- | --- | --- | --- |
| **BiLSTM** | **25.363** | **40.221** | **56.140** | **41.578** |
| BERT | 19.012 | 31.233 | 47.355 | 33.582 |
| BioBERT | 18.710 | 31.860 | 47.288 | 33.278 |
| BioClinicalBERT | 18.141 | 30.093 | 43.280 | 31.193 |

In order to test the validity of the obtained results, we have run a paired t-test to compare the BLEU scores produced by the BiLSTM model with those produced by the second-best model - BERT. After running the statistical test, we got the p-value of approximately $1.214e^{-13}$, which was far less than the threshold of 0.05 (we have used this cutoff for the p-value in our experiments). Therefore, the obtained results were statistically significant and the BiLSTM model has outperformed all other NMT models.

We have also run statistical tests to compare BERT and BioBERT models as well as BioBERT and BioClinicalBERT models. In the case of BERT and BioBERT, we have obtained a p-value of approximately 0.459, which is greater than the cutoff value of 0.05 and therefore, the result was not statistically significant. For BioBERT and BioClinicalBERT, the p-value was approximately $7.056e^{-6}$, which is a lot less than 0.05 and thus, we have concluded that the obtained results were statistically significant and the mean difference between BLEU scores produced by BioBERT and BioClinicalBERT models was non-zero.

Qualitatively, according to BLEU scores, BiLSTM has, on average, generated "high quality translations," while the rest of the models had, on average, produced "understandable to good translations." The BiLSTM model was the only one whose 75th percentile was "very high quality, adequate, and fluent translations." It should be noted that none of reported metrics correspond to "almost useless" translations. Similarly, none of the metrics achieved the BLEU score to be of "quality often better than human."

We have also compared the HIRs across all the data used in our experiments. The results are shown in Table 3. The lowest average HIR was observed in the target data in testing since it was a gold standard manually created by human experts. The source data in training had the highest average HIR. The HIRs of different model outputs were similar with the average scores of 0.025, 0.022, 0.026, and 0.024 for BiLSTM, BERT, BioBERT, and

BioClinicalBERT respectively. It should be noted that BERT had the lowest HIR among all models. Target data in training (silver standard) had the average HIR of 0.031 and the source data in testing had that of 0.095.

Regarding four health information websites in our dataset, Reddit.com and Mayoclinic.org had the lowest HIR in the source data in training. MedlinePlus.gov had the lowest HIR in the source data in testing. As for highest health illiteracy rates, all but one data source were featured, with Mayoclinic.org being the only data source that did not have the highest HIR in any data. The lowest HIR value of 0 was observed in BiLSTM, BERT, BioBERT, and BioClinicalBERT outputs for MedlinePlus.gov. The HIRs of four NMT models are comparable, with the BERT output having the lowest HIR. Please note that the HIR in our dataset does not indicate the health literacy level of patient education materials in the original websites.

Table 3. HIRs in the original corpus in training, silver standard in training, gold standard in testing, and the output generated by OpenNMT-, BERT-, BioBERT-, and BioClinicalBERT-based NMT approaches.

| Data Source | Source Data in Training | Target Data in Training (Silver Standard) | Source Data in Testing | Target Data in Testing (Gold Standard) | OpenNMT Output in Testing | BERT Output in Testing | BioBERT Output in Testing | BioClinicalBERT Output in Testing |
|---|---|---|---|---|---|---|---|---|
| Mayoclinic.org | 0.106 | 0.032 | 0.102 | 0.014 | 0.030 | 0.026 | 0.030 | 0.030 |
| MedlinePlus.gov | 0.125 | 0.033 | 0.050 | 0.000 | 0.000 | 0.000 | 0.000 | 0.000 |
| Drugs.com | 0.110 | 0.035 | 0.110 | 0.025 | 0.044 | 0.039 | 0.046 | 0.042 |
| Reddit.com | 0.092 | 0.023 | 0.117 | 0.011 | 0.025 | 0.023 | 0.026 | 0.025 |
| Average | 0.108 | 0.031 | 0.095 | 0.012 | 0.025 | 0.022 | 0.026 | 0.024 |

**Discussion**

We interpreted the outputs from four NMT models in comparison with the human translation in the gold standard dataset according to the translation quality metrics, including sentence structure, grammar, clarity, easiness to read and collocation. This medical terminology interpretation analysis was conducted by researchers with a nursing background.

We observed many promising results generated by the NMT models. For example, NMT models are able to identify the correct complicated words and simplify them into layman language. *"However, if your family has one child with hypoplastic left heart syndrome, the risk of having another with a similar condition is increased."* was translated into *"However, if your family has one child with a left heart disease, the chance of having another with a medical problem is added to a higher chance"*. Our NMT models simplified the medical term "*hypoplastic left heart syndrome*" into "*left heart disease*" and simplified "*risk*" into "*chance*" successfully without any sentence patterns or collocation issues. Moreover, NMT models outperforms in translating short and simple sentences. For instance, some simple sentences like *"The test can't detect all cancers"*, NMT models perform as well as human translation in identifying the term "detect" and translate into *"The test can't find all cancers"*.

We also observed several limitations of the proposed NMT methods in translating health illiterate language. First, compared to the human translation, we found that NMT models suffer from sentence completeness. For BERT, BioBERT and BioClinicalBERT models, most of them could not translate the complete sentence with the exact meaning from the original sentence. For example, if the sentence ends with a term that the CDC Plain Language Thesaurus did not support, our models may stop generating the translation. Overall, the results of BiLSTM model contain fewer illiterate words than the BERT-based models. Second, apart from the sentence completeness, translations for certain medical terms are not appropriate or unnecessary. For example, some commonly known medical terms such as treatment, heart attack, allergic reaction are being translated as long and plain sentences. One of the reasons could be the limitation of the CDC Plain Language Thesaurus where these terms are considered as health illiterate. These terms are being translated in the training data and thus learned by the model. Third, the machine learning models still have room for improvement for complicated terms. For instance, some medical terms like eosinophilic esophagitis, hemostasis, axillary lymphadenitis, alveolar atelectasis, chickenpox, and medication names are not included in the CDC Plain Language Thesaurus. As such, the model is unable to learn from the silver standard data and unable to translate such complicated terms. Thus, using the model we trained to identify additional illiterate words and supplement with the expert curated CDC Plain Language Thesaurus is subject to a future direction.

Finally, human translation performs much better in general sentence fluency and readability while the machine learning models may generate grammatical issues in the translation results, such as tenses, plural/singular forms, collocation, and sentence patterns. In particular, our models may have some grammatical flaws that could affect the readability for lengthy sentences. For example, *"A weak immune system caused by disease or by using certain medicines"* was translated as *"A weak protected from system suddenly overcome by a disease or by using certain part."* in machine learning model. Some medical nouns may look similar but their semantic meanings are completely different. For instance, *"Your doctor may also take biopsies of the esophagus to look for inflammation"* was translated as *"Your doctor may also take the esophagus to look for the swelling"* in human annotation, while machine learning model translated into *"Your doctor may also take the esophagus to look for the sore"*. Inflammation is not equivalent as sore in medical terminology, while swelling can be partially caused by inflammation. In addition, some terms are being translated into different meanings by both human translation and machine learning models, such as treatment was translated as "medicine" and "therapy", healthcare provider was translated as "healthcare supply" and "healthcare order". As such, interpretation of meanings could be confusing if these results were given to illiterate patients who do not have adequate knowledge to differentiate the difference.

Leveraging NLP to identify literate words into illiterate words is important to improve the health literacy among patients with low health literacy from certain populations and low socio-economic status background. Based on our findings and analysis, we highly recommend future studies should create a dataset for common medical terms, such as heart attack, hypertension, cancer and tumor, and expand the CDC Plain Language Thesaurus. These terms should not be translated by the machine learning model as they are commonly known and further translation may be confusing and misleading to patients.

In order to improve the accuracy and translation quality of the machine learning model, performing systematic language checks is necessary. Our analysis identified an array of minor grammatical and collocation issues, it is recommended that future studies should leverage the existing language assistance models to counter check the results automatically. All the final translation should be approved by healthcare experts in the field to ensure the translation quality before delivering to patients.

**Conclusion**

Health literacy is associated with health outcomes and health disparities, particularly for certain populations. Current patient education materials require high health literacy, which makes it difficult for people with low health literacy to understand health information, follow post-visit instructions, manage their chronic conditions, and use prescriptions. In this study, we trained and evaluated the state-of-the-art neural machine translation (NMT) models based on patient education materials scraped from four online health information websites: namely MedlinePlus.gov, Drugs.com, Mayoclinic.org and Reddit.com. The experimental results showed the effectiveness of NMT models in translating health illiterate language. Meanwhile, we also identified limitations in the proposed approaches and recommended a few future directions to improve health literacy in patient education materials.

**Acknowledgement**

The authors would like to acknowledge support from the University of Pittsburgh Momentum Funds, Clinical and Translational Science Institute Exploring Existing Data Resources Pilot Awards, and the School of Health and Rehabilitation Sciences Dean's Research and Development Award.

**Acknowledgement**

The authors would like to acknowledge support from the University of Pittsburgh Momentum Funds, Clinical and Translational Science Institute Exploring Existing Data Resources Pilot Awards, and the School of Health and Rehabilitation Sciences Dean's Research and Development Award.

**References**

1.  Pronk NP, Kleinman DV, Richmond TS. Healthy People 2030: Moving toward equitable health and well-being in the United States. EClinicalMedicine. 2021 Mar;33:100777. PMCID: PMC7941044

2.  Brach C, Harris LM. Healthy People 2030 Health Literacy Definition Tells Organizations: Make Information and Services Easy to Find, Understand, and Use. J Gen Intern Med. 2021 Apr;36(4):1084–1085. PMCID: PMC8042077

3.  Schillinger D. The Intersections Between Social Determinants of Health, Health Literacy, and Health Disparities. Stud Health Technol Inform. 2020 Jun 25;269:22–41. PMCID: PMC7710382

4.  Vernon JA, University of Connecticut. Department of Finance. Low Health Literacy: Implications for National Health Policy. 2007.

5.  Muvuka B, Combs RM, Ayangeakaa SD, Ali NM, Wendel ML, Jackson T. Health Literacy in African-American Communities: Barriers and Strategies. Health Lit Res Pract. 2020 Jul 16;4(3):e138–e143. PMCID: PMC7365659

6.  Brice JH, Travers D, Cowden CS, Young MD, Sanhueza A, Dunston Y. Health literacy among Spanish-speaking patients in the emergency department. J Natl Med Assoc. 2008 Nov;100(11):1326–1332. PMID: 19024230

7.  Liu YB, Chen YL, Xue HP, Hou P. Health Literacy Risk in Older Adults With and Without Mild Cognitive Impairment. Nurs Res. 2019;68(6):433–438. PMCID: PMC6845310

8.  Stossel LM, Segar N, Gliatto P, Fallar R, Karani R. Readability of patient education materials available at the point of care. J Gen Intern Med. 2012 Sep;27(9):1165–1170. PMCID: PMC3514986

9.  Mcinnes N, Haglund BJA. Readability of online health information: implications for health literacy. Inform Health Soc Care. 2011;36(4):173–189.

10. Daraz L, Morrow AS, Ponce OJ, Farah W, Katabi A, Majzoub A, Seisa MO, Benkhadra R, Alsawas M, Larry P, Murad MH. Readability of Online Health Information: A Meta-Narrative Systematic Review. Am J Med Qual. 2018 Jan 18;33(5):487–492. PMID: 29345143

11. Chandrasekar R, Doran C, Srinivas B. Motivations and methods for text simplification [Internet]. Proceedings of the 16th conference on Computational linguistics -. 1996. Available from: http://dx.doi.org/10.3115/993268.993361

12. Miller GA. WordNet: An Electronic Lexical Database. MIT Press; 1998.

13. Devlin SL. Simplifying Natural Language for Aphasic Readers. 1999.

14. Zhu Z, Bernhard D, Gurevych I. A monolingual tree-based translation model for sentence simplification. Proceedings of the 23rd International Conference on Computational Linguistics (Coling 2010). p. 1353–1361.

15. Wubben S, Van Den Bosch A, Krahmer E. Sentence simplification by monolingual machine translation. Proceedings of the 50th Annual Meeting of the Association for Computational Linguistics (Volume 1: Long Papers). p. 1015–1024.

16. Zhang X, Lapata M. Sentence Simplification with Deep Reinforcement Learning [Internet]. Proceedings of the 2017 Conference on Empirical Methods in Natural Language Processing. 2017. Available from: http://dx.doi.org/10.18653/v1/d17-1062

17. Keselman A, Logan R, Smith CA, Leroy G, Zeng-Treitler Q. Developing informatics tools and strategies for consumer-centered health communication. J Am Med Inform Assoc. 2008 Jul;15(4):473–483. PMCID: PMC2442255

18. Kandula S, Zeng-Treitler Q. Creating a Gold Standard for the Readability Measurement of Health Texts. AMIA annual symposium proceedings. p. 353.

19. Kandula S, Curtis D, Zeng-Treitler Q. A semantic and syntactic text simplification tool for health content. AMIA Annu Symp Proc. 2010 Nov 13;2010:366–370. PMCID: PMC3041424

20. Kauchak D, Leroy G, Hogue A. Measuring Text Difficulty Using Parse-Tree Frequency. J Assoc Inf Sci Technol. 2017 Sep;68(9):2088–2100. PMCID: PMC5644354

21. Klein G, Kim Y, Deng Y, Senellart J, Rush A. OpenNMT: Open-Source Toolkit for Neural Machine Translation [Internet]. Proceedings of ACL 2017, System Demonstrations. 2017. Available from: http://dx.doi.org/10.18653/v1/p17-4012

22. Vaswani A, Shazeer N, Parmar N, Uszkoreit J, Jones L, Gomez AN, Kaiser Ł, Polosukhin I. Attention is all you need. Advances in neural information processing systems. 2017.

23. Zhao H, Lu J, Cao J. A short text conversation generation model combining BERT and context attention mechanism [Internet]. International Journal of Computational Science and Engineering. 2020. p. 136. Available from: http://dx.doi.org/10.1504/ijcse.2020.110536

24. Fan X, Feng X, Dong Y, Hou H. COVID-19 CT image recognition algorithm based on transformer and CNN. Displays. 2022 Apr;72:102150. PMCID: PMC8785369

25. Devlin J, Chang MW, Lee K, Toutanova K. Bert: Pre-training of deep bidirectional transformers for language understanding. Proceedings of NAACL-HLT. 2019. p. 4171–4186.

26. Lee J, Yoon W, Kim S, Kim D, Kim S, So CH, Kang J. BioBERT: a pre-trained biomedical language representation model for biomedical text mining. Bioinformatics. 2020 Feb 15;36(4):1234–1240. PMCID: PMC7703786

27. Alsentzer E, Murphy J, Boag W, Weng WH, Jindi D, Naumann T, McDermott M. Publicly available clinical BERT embeddings. Proceedings of the 2nd Clinical Natural Language Processing Workshop. 2019. p. 72–78.

28. Papineni K, Roukos S, Ward T, Zhu WJ. BLEU: a method for automatic evaluation of machine translation. Proceedings of the 40th annual meeting of the Association for Computational Linguistics. 2002. p. 311–318.